\documentclass{article}
\usepackage[affil-it]{authblk}
\usepackage{color}
\usepackage[ruled,vlined]{algorithm2e}
\usepackage{amsmath,mathtools, amssymb,amsfonts,amsthm}
\usepackage[inline]{enumitem}
\setlist[itemize]{leftmargin=20mm}
\usepackage{graphicx}
\usepackage{booktabs,tabularx}
\usepackage{siunitx}
\usepackage{multirow}
\usepackage{pgf,tikz}
\usepackage{pgfplots}
\usetikzlibrary{shapes, pgfplots.fillbetween} 
\usepackage{url}
\usepackage{commath}
\usepackage{caption}
\usepackage{subcaption}
\usepackage{gensymb}
\usepackage[hidelinks]{hyperref}
\usepackage{cleveref}
\newcommand{\mosinn}[1]{{\color{black}{#1}}}
\newcommand{\mosir}[1]{{\color{black}{#1}}}

\newcommand{\mosi}[1]{{\color{black}{#1}}}

\newcommand{\tomas}[1]{{\color{black}{#1}}}

\title{Automated machine learning for borehole resistivity measurements}

\author[1]{M. Shahriari}
\author[2,3,4]{D. Pardo}
\author[1]{S. Kargaran}
\author[2]{T. Teijeiro} 
	
\affil[1]{\footnotesize Software Competence Center Hagenberg GmbH (SCCH), Hagenberg, Austria}
\affil[2]{\footnotesize University of the Basque Country (UPV/EHU), Leioa, Spain}
\affil[3]{\footnotesize Basque Center for Applied Mathematics, (BCAM), Bilbao, Spain}
\affil[4]{\footnotesize Ikerbasque (Basque Foundation for Sciences), Bilbao, Spain}
\date{\today}

\begin{document}
\maketitle
\begin{abstract}
Deep neural networks (DNNs) offer a real-time solution for the inversion of borehole resistivity measurements to approximate forward and inverse operators. Using a extremely large DNN to approximate the operators is possible, but it demands a considerable training time. Moreover, evaluating the network after training also requires a significant amount of memory and processing power. In addition, we may overfit the model. In this work, we propose a scoring function that accounts for the accuracy and size of the DNNs compared to a reference DNN that provides a good approximation for the operators. Using this scoring function, we use DNN architecture search algorithms to obtain a quasi-optimal DNN smaller than the reference network; hence, it requires less computational effort during training and evaluation. The quasi-optimal DNN delivers comparable accuracy to the original large DNN.

\end{abstract}

\section{Introduction}
Oil and gas companies employ geosteering to increase the productivity of their wells \cite{Bittar1, Beer}. \mosi{In this application, a logging-while-drilling (LWD) instrument helps us to navigate the well trajectory inside the oil reservoir to maximize its production.} A LWD instrument incorporates transmitters and receivers, in our case, electromagnetic (EM) ones \cite{spies1996electrical, samouelian2005electrical, Shahriari, Desbrandes}.

There are two types of mathematical problems in geosteering: the forward and the inverse. In the forward problem, for a given earth subsurface and trajectory, we simulate measurements at the receivers by solving a partial differential equation (PDE) with boundary conditions. In the case of EM measurements, we solve Maxwell's equations assuming a zero Dirichlet boundary condition far away from the transmitters \cite{Shahriari, Davydycheva, Loseth}. In the inverse problem, given the recorded measurements at the receivers, we estimate the subsurface properties by minimizing a loss function \cite{Tarantola}. 

Some traditional methods to solve the inverse problem include gradient-based and statistics-based approaches \cite{Tarantola, Watzenig,Pardo,Ijasan, Malinverno_2000}. Artificial intelligence (AI) algorithms, particularly deep learning (DL), have recently become popular to solve the inverse problem \cite{Shahriari_deep_inverse, Shahriari_loss,jin2019using, Puzyrev, Moghadas, 7949028, Hu}. In this work, we employ a deep neural network (DNN) to approximate the solution of the inverse problem.

The main difficulty when solving the inverse problem arises because of the non-uniqueness of its solution, i.e., there exist multiple outputs for each input~\cite{Tarantola}. It turns out that the DNN approximation may become an average of all the existing solutions, which can be far from any of them. In \cite{Shahriari_loss}, we proposed a specific loss function based on the misfit of the measurements that incorporates both the inverse and forward solutions. To reduce the computational time, we approximated the forward function (the solution of the PDEs) using a DNN~\cite{Shahriari_deep_forward, Alyaev}. We used a two-step training to approximate the inverse operator. In the first step, we approximated the forward function. In the second step, using the trained DNN approximation of the forward function and the introduced loss function, we approximated the solution of the inverse operator. This approach guaranteed that the output of the trained DNN approximation of the inverse operator delivers one of its solutions. However, \cite{Shahriari_loss} does not discuss the optimal selection of the DNN architecture, resulting in a large DNN to achieve its goal. Here, we discuss a proper selection of the DNN architecture to approximate both the forward and inverse operators involved in the aforementioned two-step training.

Designing DNN architectures by hand is difficult \cite{Lu, Higham, goodfellow2016deep}. An excessively large DNN may achieve the required accuracy, but we may incur in excessive computational costs and possibly in overfitting. On the other side, using a small DNN may limit the accuracy. \mosi{Here, we employ automated machine learning (AutoML) algorithms, specifically DNN architecture search algorithms, to build quasi-optimal DNN architectures that balance size and accuracy of the networks \cite{hutter2019automated,he2021automl,omalley2019kerastuner,jin2019auto, Elsken2019NeuralAS}. These search techniques allow us to find well-suited DNNs with limited knowledge about the DNN architectures.}

\tomas{In this work, we have considered as a reference model the DNN described in~\cite{Shahriari_tool}, and we have evaluated the explored architectures by assessing the error variation and the number of trainable parameters. This is tightly related to model compression~\cite{Cheng2018}, a research field that explores methods for reducing the complexity of a reference model without affecting its accuracy. In particular, our approach can be linked to knowledge distillation~\cite{Caruana2014}, a family of techniques consisting of training a more compact model that integrates the output of the reference model in the loss function. In our case, we rely on AutoML for exploring the space of compact models. }\mosinn{The resulting DNN architecture is a parametric representation of the desired operator, i.e., forward and inverse. Hence, having a smaller model makes it computationally cheaper to train a DNN when we need to train the model on new datasets (new scenarios). Moreover, smaller DNNs require less memory to save and fewer computational resources to evaluate. Therefore, saving and evaluating on portable devices, e.g., LWD instruments, is more versatile.}

\tomas{Another popular and simpler method for model compression is parameter pruning~\cite{Srinivas2015}, consisting of removing redundant and uncritical parameters from a model after it has been trained (e.g., by setting to zero all the weights that have a value below a threshold). Thus, the main utility of pruning is to simplify model evaluation. In our case, since the main objective is to find a simpler architecture that can be efficiently trained on new scenarios, we do not consider parameter pruning as a suitable approach.}

We consider as our main architectural component a convolutional block composed of three one-dimensional convolutional layers \cite{He}. The final DNN consists of a specific number of the mentioned blocks placed sequentially, one after another. Hence, to obtain the quasi-optimal DNN, we find the associated parameters to the convolutional block and the number of blocks, i.e., the set of hyperparameters associated with the aforementioned DNN architecture. For this, we introduce a scoring function that accounts both for the loss and size of the DNN, and by minimizing this scoring function, we find the quasi-optimal DNN. We employ two standard architecture search algorithms; namely, random and Bayesian searches~\cite{Elsken2019NeuralAS}. We compare the results of the obtained DNN vs. our original DNN designed by hand. Results show that the AutoML DNN algorithms deliver smaller DNN networks that preserve the accuracy of the larger DNN created by hand. \mosir{Throughout this work, we consider noise-free synthetic measurements. Nonetheless, we expect smaller DNNs to be more resilient towards noise than large DNNs that are more prone to overfitting.}

The remaining of this work is organized as follows: \Cref{sec:definition} defines the forward and inverse problems in the inversion of borehole resistivity measurements. \Cref{sec:training} describes a two-step training strategy that we use in this work to obtain the DNN approximation of the inverse operator. \Cref{sec:architecture} discusses the considered DNN architecture components, the definition of the scoring function, and the DNN architecture search algorithms that we use in this work. \Cref{sec:results} verifies the proposed techniques by showing the training results and the model's output for some synthetic models. \Cref{sec:conclusion} is dedicated to the conclusion.

\section{Problem definition}
\label{sec:definition}

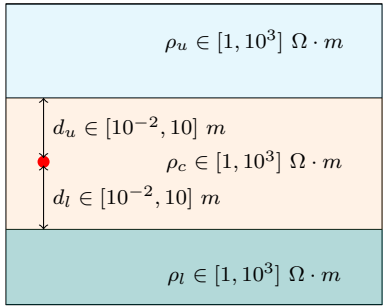
\begin{figure}[h]
	\centering
	\begin{tikzpicture}[scale=1.0]
	\draw (2,0) rectangle (7,4);
	\draw[name path=A1] (2,0) -- (7,0);
	\draw[name path=A2] (2,4) -- (7,4);
	\draw[name path=line1] (2,2.75) -- (7,2.75);
	\draw[name path=line2] (2,1) -- (7,1);
	\tikzfillbetween[of=line2 and A1]{teal, opacity=0.3};
	\tikzfillbetween[of= line2 and line1]{orange, opacity=0.1};
	\tikzfillbetween[of= line1 and A2]{cyan, opacity=0.1};	
	\fill[red] (2.5, 1.9) circle (0.08cm);
	\draw[<->] (2.5, 1.95) -- (2.5,2.75) node [midway, right] {{\footnotesize {$d_{u} \in [10^{-2},10]\ m $}}};
	\draw[<->] (2.5,1.85) -- (2.5,1.0) node [midway, right] {{\footnotesize {$d_{l} \in [10^{-2},10]\ m $}}};
	\node (rho_) at (5.3,1.9) {{\footnotesize {$\rho_{c} \in [1,10^3]\ \Omega \cdot m $}}};
	\node (rho_u) at (5.3,3.5) {{\footnotesize {$\rho_{u} \in [1,10^3]\ \Omega \cdot m $}}};
	\node (rho_l) at (5.3,0.4) {{\footnotesize {$\rho_{l} \in [1,10^3]\ \Omega \cdot m $}}};
	\end{tikzpicture}
	\caption{1D subsurface formation, its parameterization and the range of variation of the parameters. The logging position indicated by the red circle. $\rho_u$, $\rho_c$, and $\rho_l$ are the resistivities of the upper, central, and lower layers, respectively. $d_u$ and $d_l$ are vertical distances from the current logging position to the upper and lower bed boundaries, respectively.}
	\label{fig:1D}
\end{figure}

Let $\mathbf{p}$ be the subsurface properties. In this application, since the inversion should be performed in real-time, to reduce the computational complexity of the problem it is common to consider a 1D-layered formation around the logging position \cite{Pardo, Shahriari, Davydycheva}. Hence, $\mathbf{p}$ is a vector of variables parameterizing the 1D formation as follows:

\begin{equation}
\mathbf{p} = (\rho_c, \rho_u, \rho_l, d_u, d_l),
\end{equation}
where $\rho_c$ is the resistivity of the host (central) layer of the logging instrument, $\rho_u$ and $\rho_l$ are the resistivities of the upper and lower layers, respectively, and $d_u$ and $d_l$ are the vertical distances to the upper and lower bed boundaries, respectively. \Cref{fig:1D} shows the earth subsurface parameterization and corresponding variation intervals selected based on their occurrence on the geological targets~\cite{Pardo, Shahriari_deep_inverse, Shahriari_owt}.

\begin{figure*}[ht]
	\centering
	\begin{tikzpicture}
	\node at (0,3.5)[scale=0.8]{\begin{tikzpicture}

\fill[gray!80!white] (-1.3*5,0) -- (1.3*5,0.) --  
 (1.3*5, 0.15*3) -- (-1.3*5,0.15*3) -- cycle;


\fill[black!80!white] (-0.4064*5-0.02*3,0) -- (-0.4064*5+0.02*3,0.) node[below] {\footnotesize \bf \textcolor{black}{Tx\textsubscript{1,1}}} -- (-0.4064*5+0.02*3, 0.15*3)  -- (-0.4064*5-0.02*3,0.15*3) -- cycle;

\fill[black!80!white] (0.4064*5-0.02*3,0) -- (0.4064*5+0.02*3,0.) node[below] {\footnotesize \bf \textcolor{black}{Tx\textsubscript{1,2}}}  -- (0.4064*5+0.02*3, 0.15*3)  -- (0.4064*5-0.02*3,0.15*3) -- cycle;


\fill[black!80!white] (-0.4064*15-0.02*3,0) -- (-0.4064*15+0.02*3,0.) node[below] {\footnotesize \bf \textcolor{black}{Tx\textsubscript{2,1}}} -- (-0.4064*15+0.02*3, 0.15*3)  -- (-0.4064*15-0.02*3,0.15*3) -- cycle;

\fill[black!80!white] (0.4064*15-0.02*3,0) -- (0.4064*15+0.02*3,0.) node[below] {\footnotesize \bf \textcolor{black}{Tx\textsubscript{2,2}}} -- (0.4064*15+0.02*3, 0.15*3)  -- (0.4064*15-0.02*3,0.15*3) -- cycle;

\fill[red!80!white] (-0.1016*5-0.02*3,0) -- (-0.1016*5+0.02*3,0.)  -- (-0.1016*5+0.02*3, 0.15*3) node[above] {\footnotesize \bf \textcolor{black}{Rx$_1$}} -- (-0.1016*5-0.02*3,0.15*3) -- cycle;
\fill[red!80!white] ( 0.1016*5-0.02*3,0) -- ( 0.1016*5+0.02*3,0.) -- ( 0.1016*5+0.02*3, 0.15*3)  node[above] {\footnotesize \bf \textcolor{black}{Rx$_2$}} -- ( 0.1016*5-0.02*3,0.15*3) -- cycle;

\draw[black, line width=1pt,<->] (-0.1016*5, -0.6)      -- (0.1016*5, -0.6) node[pos=0.5, below] {\footnotesize \bf \textcolor{black}{\SI{0.2032}{\m}}}  ;
\draw[gray, dashed] (-0.1016*5, -0.6)      -- (-0.1016*5, 0)  ;
\draw[gray, dashed] ( 0.1016*5, -0.6)      -- ( 0.1016*5, 0)  ;

\draw[black, line width=1pt,<->] (-0.1016*20, 1.05)      -- (0.1016*20, 1.05) node[pos=0.5, above] {\footnotesize \bf \textcolor{black}{\SI{0.8128}{\m},~\SI{2}{\mega \hertz}}};
\draw[gray, dashed] (-0.1016*20, 1.05)      -- (-0.1016*20, 0.45)  ;
\draw[gray, dashed] ( 0.1016*20, 1.05)      -- ( 0.1016*20, 0.45)  ;


\draw[black, line width=1pt,<->] (-0.1016*60, 1.65)      -- (0.1016*60, 1.65) node[pos=0.5, above] {\footnotesize \bf \textcolor{black}{\SI{2.4384}{\m},~\SI{0.25}{\mega \hertz}}};
\draw[gray, dashed] (-0.1016*60, 1.65)      -- (-0.1016*60, 0.45)  ;
\draw[gray, dashed] ( 0.1016*60, 1.65)      -- ( 0.1016*60, 0.45)  ;

\node[rotate=90] (I_h) at (-7,0.30) {Conventional LWD};

\end{tikzpicture}};
	\node at (0,0.0)[scale=0.8]{\begin{tikzpicture}

\fill[gray!80!white] (-1.3*5,0) -- (1.3*5,0.) --  
(1.3*5, 0.15*3) -- (-1.3*5,0.15*3) -- cycle;


\fill[black!80!white] (-0.6096*10-0.02*3,0) -- (-0.6096*10+0.02*3,0.)
node[below] {\footnotesize \bf \textcolor{black}{Tx}} -- (-0.6096*10+0.02*3, 0.15*3)   -- (-0.6096*10-0.02*3,0.15*3) -- cycle;
%
\fill[red!80!white] (-0.02*3,0) -- (0.02*3,0.)  -- (0.02*3, 0.15*3) node[above] {\footnotesize \bf \textcolor{black}{Rx\textsubscript{1}}} -- (-0.02*3,0.15*3) -- cycle;

\fill[red!80!white] (0.6096*10-0.02*3,0) -- (0.6096*10+0.02*3,0.)node[below] {\footnotesize \bf \textcolor{black}{Rx\textsubscript{2}}}  -- (0.6096*10+0.02*3, 0.15*3)  -- (0.6096*10-0.02*3,0.15*3) -- cycle;

\draw[black, line width=1pt,<->] (-0.6096*10, -0.6)      -- (0, -0.6) node[pos=0.5, below] {\footnotesize \bf \textcolor{black}{ \SI{12}{\m},~\SI{24}{\kilo \hertz}}}  ;
\draw[gray, dashed] (-0.6096*10,-0.6)      -- (-0.6096*10, 0)  ;
\draw[gray, dashed] ( 0.0,-0.6)      -- ( 0.0, 0)  ;

\draw[black, line width=1pt,<->] (-0.6096*10, 1.05)      -- (0.6096*10, 1.05) node[pos=0.5, above] {\footnotesize \bf \textcolor{black}{\SI{25}{\m},~\SI{2}{\kilo \hertz}}}  ;
\draw[gray, dashed] (-0.6096*10,1.05)      -- (-0.6096*10, 0.45)  ;
\draw[gray, dashed] ( 0.6096*10,1.05)      -- ( 0.6096*10, 0.45)  ;

\node[rotate=90] (I_h) at (-7.,0.30) {Deep azimuthal};

\end{tikzpicture}};
	\end{tikzpicture}
	\caption{LWD instruments. \textbf{Tx}, and \textbf{Tx\textsubscript{i,j}}, \textbf{i}, \textbf{j}=1,2, denote the transmitters. \textbf{Rx\textsubscript{1}}, \textbf{Rx\textsubscript{2}} are the receivers.}
	\label{fig:lwd_instruments}
\end{figure*}
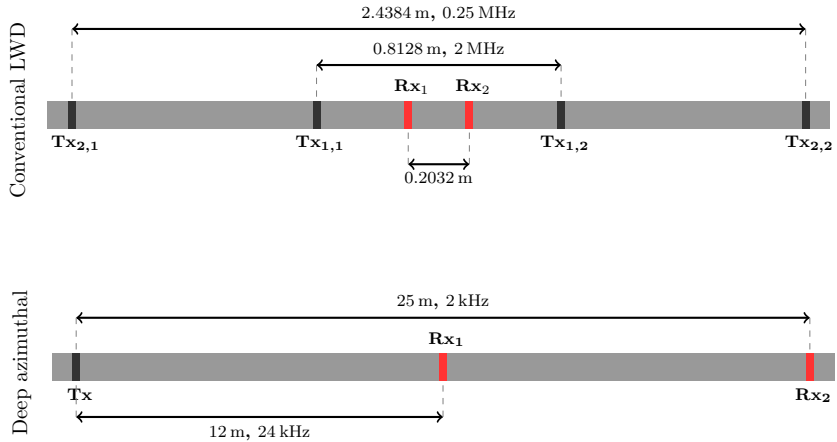

To evaluate the measurements, we use two logging instruments: a conventional LWD and an azimuthal one (see \Cref{fig:lwd_instruments}). We consider ${\mathbf{m}}$ to be the measurements obtained at the receivers using the aforementioned logging instruments. \mosi{\Cref{tab:measurement_def} defines those measurements. \Cref{tab:tools_measurements} shows the evaluated measurements for each transmitter-receiver set. For each measurement, we obtain a real and an imaginary part, except for the geosignal as the imaginary part is discontinuous (see \cite{Shahriari_tool} for more details regarding the selection of these measurements). Therefore, ${\mathbf{m}}$ is a set of 13 measurements. Moreover, we consider high-angle (almost horizontal) trajectories, hence, for the case of trajectory dip angle, we have $t \in [83\degree, 97\degree]$. Then, we have the following separate problems:}


\begin{itemize}
	\item Forward problem: Given $\mathbf{p}$ and $t$, we obtain $\bf{m}$ at the receivers, i.e., ${\cal F} ({\bf p}, t) = \bf{m}$, where ${\cal F}$ is the solution of Maxwell's equations with a zero Dirichlet boundary condition far away from the transmitters~\cite{Shahriari, Shahriari_owt, Davydycheva, Davydycheva1, Shahriari_deep_forward, Alyaev}.
	
	\item Inverse problem: Given the measurements acquired at the receivers and the trajectory dip angle, the inverse operator ${\cal I}$ delivers the subsurface properties, i.e., ${\cal I} ({\bf m}, t) = \bf{p}$ \cite{Pardo, Shahriari_deep_inverse, Shahriari_loss, jin2019using, Ijasan}.
\end{itemize}

In this work, we use DNNs to approximate the forward function ${\cal F}$ and inverse operator ${\cal I}$. Training a DNN requires a large dataset. Hence, given the above subsurface parameterization, trajectory, and measurements, we produce a dataset of 300,000 randomly selected samples using a fast semi-analytic solver~\cite{Loseth}. We then express the values of the subsurface properties in the logarithmic scale \cite{Shahriari_loss}, and rescale all the variables (i.e., subsurface properties in the logarithmic scale and the measurements) to the interval $[0.5,1.5]$ (see \cite{Shahriari_tool} for details).

\begin{table}[h]
	\begin{center}
		\begin{tabular}{|l|l|l|l}
			\toprule
			Name & Measurement definition\\
			\midrule
			zz & $H_{zz}$ \\
			yy & $H_{yy}$ \\
			Geosignal & $\displaystyle\frac{H_{zz}-H_{zx}}{H_{zz}+H_{zx}}$  \\
			Symmetrized directional & $\displaystyle\frac{H_{zz}+H_{zx}}{H_{zz}-H_{zx}}\cdot \frac{H_{zz}-H_{xz}}{H_{zz}+H_{xz}}$ \\
			\bottomrule
		\end{tabular}
	\end{center}
	\caption{ Evaluated measurements and their definitions. $H_{ij}$ is the complex-valued magnetic field, where $i$ and $j$ indicate the orientations of transmitters and receivers, respectively.}
	\label{tab:measurement_def}
\end{table}

\begin{table}[h]
	\begin{center}
		\begin{tabular}{|l|l|l|l}
			\toprule
			Transmitter-receiver & Measured component\\
			\midrule
			$(Tx_{1,1}, Tx_{1,2}, Rx_1, Rx_2)$ & zz, yy, Geosignal, Symmetrized directional\\
			$(Tx_{2,1}, Tx_{2,2}, Rx_1, Rx_2)$ & Symmetrized directional\\
			$(Tx, Rx_1)$ & zz\\
			$(Tx, Rx_2)$ & Symmetrized directional  \\
			\bottomrule
		\end{tabular}
	\end{center}
	\caption{Evaluated measurements for each transmitter-receiver set.}
	\label{tab:tools_measurements}
\end{table}

\section{Two-step training strategy}
\label{sec:training}
Due to the non-uniqueness of the inverse operator's output, using conventional loss functions (based on the misfit of the inversion variables ${\mathbf p}$) may produce an inaccurate solution \cite{Shahriari_loss,Tarantola}. To guarantee that the trained DNN delivers one of the true solutions of the inverse operator, we use a two-step training strategy, as described in \cite{Shahriari_loss}. First, we approximate the forward function ${\cal F}$ as:

\begin{equation}
{\cal F}_{\alpha ^\ast} := \arg \min_{\alpha}  \sum_{i=1}^{n_{t}} L({\cal F}_{\alpha} (t_i,\mathbf{p}_i), \mathbf{m}_i),\\
\label{eq:first_step_loss}
\end{equation}
where, for given vectors ${\bf x}$ and ${\bf y}$, we define $L(\mathbf{x} , \mathbf{y})=\| \mathbf{x} - \mathbf{y} \|_{l_1}$, $\{({\mathbf p}_i,{\mathbf m}_i, t_i )\}^{n_t}_{1}$ is the training dataset consisting of $n_t$ samples, and $\alpha$ is the set of weights and biases corresponding to the DNN. Then, using the trained DNN approximation of ${\cal F}$, we use the following loss function based on the misfit of the measurements to obtain the DNN approximation of the inverse operator:

\begin{equation}
\begin{split}
{\cal I}_{\beta ^\ast} := \arg \min_{\beta} \sum_{i=1}^{n_{t}} L({\cal F}_{\alpha ^\ast} \circ {\cal I}_{\beta} (t_i,\mathbf{m}_i) , \mathbf{m}_i),
\end{split}
\label{eq:second_step_loss}
\end{equation}
where $\beta$ represents the weights and biases corresponding to the DNN, and $\circ$ indicates the composition of the functions.
\section{DNN architecture optimization}
\label{sec:architecture}
\subsection{Space of DNN architectures}
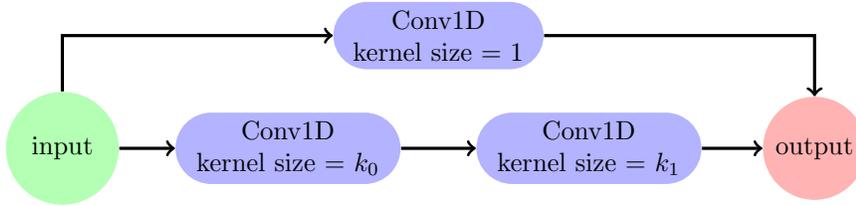
\begin{figure}
	\begin{tikzpicture}
\node[fill = green!30!white, circle, minimum size=1.5cm,align=center] (input) at (-3,0){input};

\node[fill = blue!30!white,rounded rectangle,    minimum width=1cm,
minimum height=0.8cm,align=center] (conv0) at (0,0){Conv1D \\ kernel size = $k_0$};

\node[fill = blue!30!white,rounded rectangle,    minimum width=1cm,
minimum height=0.8cm,align=center] (conv1) at (4,0){Conv1D \\ kernel size = $k_1$};

\node[fill = blue!30!white,rounded rectangle,    minimum width=1cm,
minimum height=0.8cm,align=center] (iden) at (2,1.5){Conv1D \\ kernel size = $1$};

\node[fill = red!30!white, circle, minimum size=0.3cm,align=center] (output) at (7,0){output};

\draw[->, very thick] (input) -- (conv0);

\draw[->, very thick] (conv0) -- (conv1);

\draw[->, very thick] (conv1) -- (output);

\draw[->, very thick] (iden) -- (7,1.5) --(output);

\draw[->, very thick] (input) -- (-3,1.5) --(iden);
\end{tikzpicture}
	\caption{Our convolutional block $B_{k_0, k_1}$ consists of three convolutional layers. $k_0$ and $k_1$ are kernel sizes of the convolutional layers that can vary.}
	\label{fig:resblock}
\end{figure}

We define the convolutional block $B_{k_0, k_1}$ shown in \Cref{fig:resblock} as our main architectural component of our DNNs, where $k_0$ and $k_1$ are the kernel sizes of two one-dimensional convolutional layers \cite{He}. We consider ${\cal F}_{h_f, \alpha}$ to be the DNN approximation of ${\cal F}$ given the set of hyperparameters $h_f$. Then, for $h_f=\{n, k_0, k_1, l\}$, we define ${\cal F}_{h_f, \alpha}$ as:

\begin{equation}
{\cal F}_{h_f, \alpha} = B^n_{k_0, k_1} \circ \cdots \circ B^0_{k_0, k_1} \circ C_l,
\label{eq:dnn_forward}
\end{equation}
where $n$ is the number of residual blocks and $C_l$ is a one-dimensional convolutional layer with $l$ being its kernel size. We define the search space of hyperparameters as:

\begin{equation}
{\cal S}_{\cal F}=\{n \in \{1,2,3,4\}, k_0, k_1, l \in \{3,5,7\}\}.
\end{equation}

Analogously, we consider the DNN approximation of the inverse operator ${\cal I}_{h_i, \beta}$ for a given set of hyperparameters $h_i=\{n, k_0, k_1\}$ to be as follows:
\begin{equation}
{\cal I}_{h_i, \beta} = B^n_{k_0, k_1} \circ \cdots \circ B^0_{k_0, k_1} \circ d,
\label{eq:dnn_inverse}
\end{equation}
where $d$ is a fully-connected layer with its number of nodes being the size of the vector of subsurface properties $| \mathbf{p} | = 5$. Therefore, our search space is:

\begin{equation}
{\cal S}_{\cal I}=\{n \in \{1,2,3,4, 5\}, k_0, k_1 \in \{3,5,7\}\}.
\end{equation}

\subsection{DNN hyperparameter tuning}
This work aims to find DNN approximations of ${\cal F}$ and ${\cal I}$ such that their corresponding architectures employ a minimum number of unknowns (weights and biases) and provide comparable (or better) accuracy than the excessively large reference DNN employed in \cite{Shahriari_tool}, that corresponds to the hyperparameters $h_f^o$ and $h_i^o$ for the DNN approximations of ${\cal F}$ and ${\cal I}$, respectively. For a set of hyperparameters $h_f \in {\cal S}_{\cal F}$ --see \Cref{eq:first_step_loss}-- we train its corresponding DNN defined by \Cref{eq:dnn_forward} to obtain ${\cal F}_{h_f, \alpha ^\ast}$. Then, we compute the following scoring function:

\begin{equation}
R_f(h_f) =  \underbrace{\dfrac{{\cal H}_f(h_f) - {\cal H}_f(h^o_f)}{{\cal H}_f(h^o_f)}}_{\text{relative error}}- \underbrace{\dfrac{N_p(h^o_f) - N_p(h_f)}{N_p(h^o_f)}}_{\substack{\text{relative decrease in} \\ \text{the number of unknowns}}},
\label{eq:score_forward}
\end{equation}
where
\begin{equation}
{\cal H}_f (h_f)=   \sum_{i=1}^{n_{v}} L({\cal F}_{h_f,\alpha ^\ast} (t_i,\mathbf{p}_i), \mathbf{m}_i)
\end{equation}
for $\{({\mathbf p}_i,{\mathbf m}_i, t_i )\}^{n_v}_{1}$ being a validation dataset distinct from the training dataset with $n_v$ being its size, and $N_p(h)$ is the number of unknowns of the DNN corresponding to the hyperparameter $h$. Then, the hyperparameter tuning consists of  solving the following minimization problem:
\begin{equation}
	h_f^\ast = \underset{h_f \in {\cal S}_{\cal F}}{\arg}   \min R_f(h_f).
	\label{eq:opt_forward}
\end{equation}

According to the two-step training strategy, after obtaining ${\cal F}_{h^\ast_f,\alpha ^\ast}$, we need to minimize the following problem for the hyperparameter tuning of the inverse operator:
\begin{equation}
	h_i^\ast = \underset{h_i \in {\cal S}_{\cal I}}{\arg}  \min R_i(h_i),
	\label{eq:opt_inverse}
\end{equation}
where:
\begin{equation}
R_i(h_i) = \dfrac{{\cal H}_i(h_i) - {\cal H}_i(h^o_i)}{{\cal H}_i(h^o_i)}- \dfrac{N_p(h^o_i) - N_p(h_i)}{N_p(h^o_i)},
\label{eq:score_inverse}
\end{equation}
and
\begin{equation}
{\cal H}_i (h_i) =\sum_{i=1}^{n_{v}} L({\cal F}_{h^\ast_f,\alpha ^\ast} \circ {\cal I}_{h_i,\beta^\ast} (t_i,\mathbf{m}_i) , \mathbf{m}_i).
\end{equation}

The above optimization problems have no explicit gradient formulations. The simplest method to solve these problems is a grid search, which evaluates the scoring function over all the possible combinations of the hyperparameters. However, as the evaluation of the scoring function requires a complete training of a DNN and it can be costly, \mosir{it is a common practice to rely on random search and a Bayesian approach to speed-up the optimization.} These approaches are detailed in the next section.

\subsection{AutoML algorithms}
In this section, for simplicity in the notation, we denote ${\cal S} = \{h_0, h_1, \cdots, h_n\}$ as the search space of hyperparameters and ${\cal H}$ as the scoring function. 
 
\subsubsection{Random search}
In this iterative approach, at the $ i $-th iteration, we randomly select $h_i \in {\cal S}$ as our set of hyperparameters. By training the corresponding DNN to the selected set of hyperparameters, we compute ${\cal H}(h_i)$. Generally speaking, in the case of a massive search space, it is possible to interrupt the search algorithm as soon as we achieve our goal, for instance, a specific accuracy. In our case, we repeat this process until the search space is exhausted \cite{random_search,Elsken2019NeuralAS}. We consider a search space exhausted when in five consecutive iterations, the randomly selected set of hyperparameters are amongst the ones we have already tried.

Using a random search approach to tune the hyperparameters could become excessively costly as we need to compute the score, i.e., to train a new DNN at each iteration. Moreover, we do not use the information we obtain during the previous iterations to select the next hyperparameter. As a result of such a blind selection process, this approach imposes a high computational cost, especially when considering a massive search space. Furthermore, as the selection is entirely random, and we may not try all the search space, there is no guarantee that we obtain the quasi-optimal set of hyperparameters. \mosir{However, if stopping criteria while tuning is imposed, e.g., the tuning stops when we achieve a specific value of the scoring function, it is possible that a random search obtains the hyperparameters sooner than a grid search, hence, the possibility of reduced computational cost. In the worst-case scenario, random and grid searches impose the same computational cost.}

\subsubsection{Bayesian approach}

Random search constitutes an improvement over grid search in terms of performance. However, it requires a large number of samples to properly characterize the search space. Given the high cost of calculating the scoring function for each sample, we consider a surrogate model --also known as \textit{performance predictor}~\cite{White2021}-- to: 1) estimate the scoring function without having to train the associated DNN, and 2) to select the most promising set of hyperparameters to test. For this, we use a probabilistic performance predictor based on Gaussian Processes (GPs)~\cite{Bayesian, Bayesian_Nas, Elsken2019NeuralAS}.

\paragraph{a) Gaussian Processes as performance predictors:} GPs are a generalization of multivariate Gaussian distributions to infinite dimensions, and therefore they can model a probability distribution over continuous functions. \tomas{Thus, they allow us to estimate the value of a function and its uncertainty at any point in the domain}. In our case, the function to model is \tomas{the scoring function} $\mathcal{H} : \mathcal{S} \rightarrow \mathbb{R}$.

A GP assumes that any finite set of $n$ points has an associated $n$-variate Gaussian distribution, which is completely determined by its mean vector $\mu$ and covariance matrix $\Sigma$. Since a GP is a model on a potentially infinite set of points, it is characterized by a mean function $m(h)=\mathbb{E}(\mathcal{H}(h))$ and a covariance function (a.k.a kernel) $k(h, h')=\mathbb{E}[(\mathcal{H}(h)-m(h))(\mathcal{H}(h')-m(h'))]$ \tomas{, where $\mathbb{E}(X)$ is the expected value of $X$}. These functions can be used to derive $\mu$ and $\Sigma$ for any set of points $\{h_0,\ldots,h_n\}$. All the relevant properties of the GP including continuity, differentiability, and periodicity are determined by the covariance function.

\begin{figure}[t]
     \centering
     \begin{subfigure}[t]{0.49\textwidth}
         \centering
         \includegraphics[width=\textwidth]{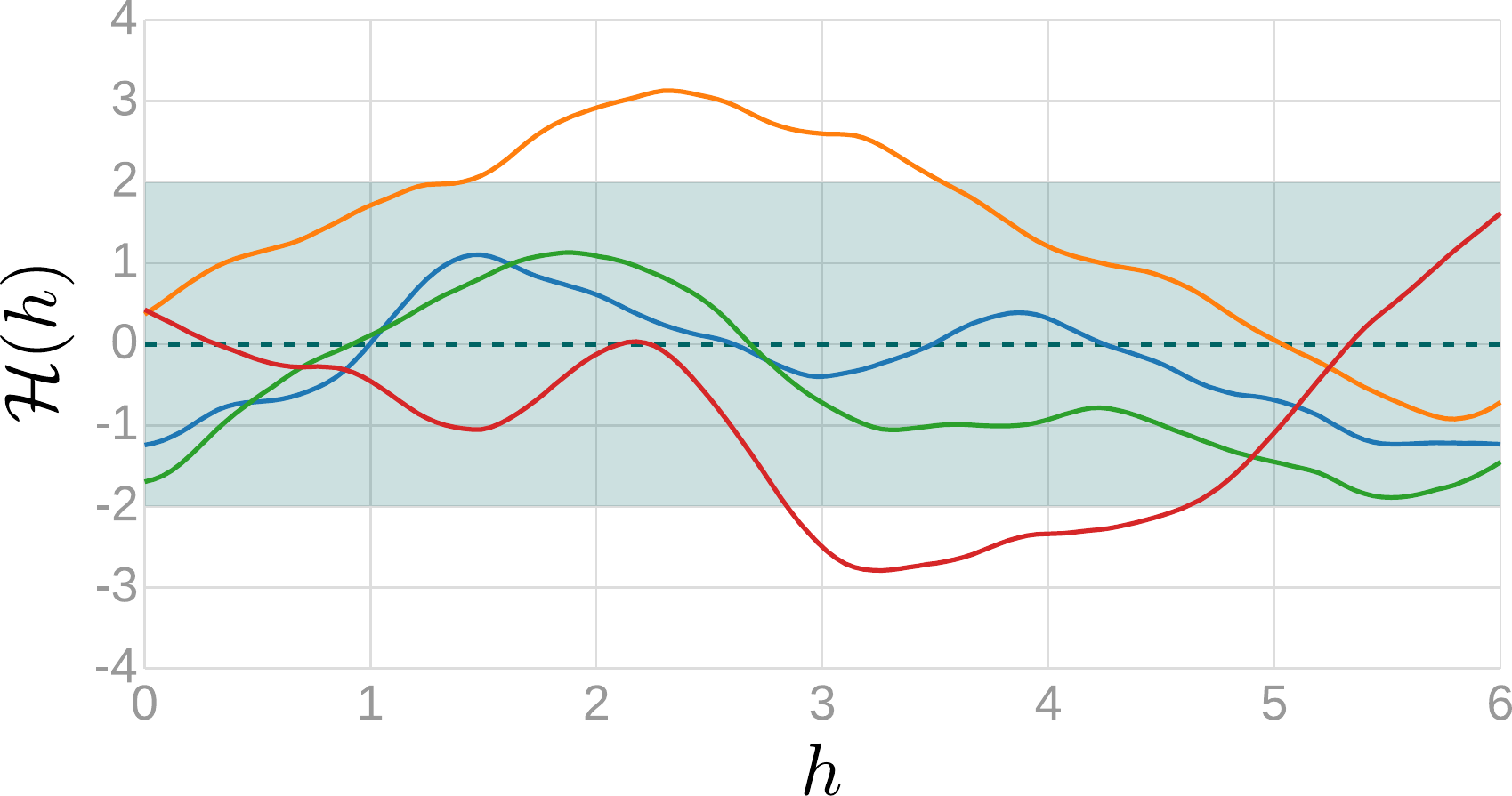}
         \caption{Prior distribution.}
         \label{fig:gp_prior}
     \end{subfigure}
     \hfill
     \begin{subfigure}[t]{0.49\textwidth}
         \centering
         \includegraphics[width=\textwidth]{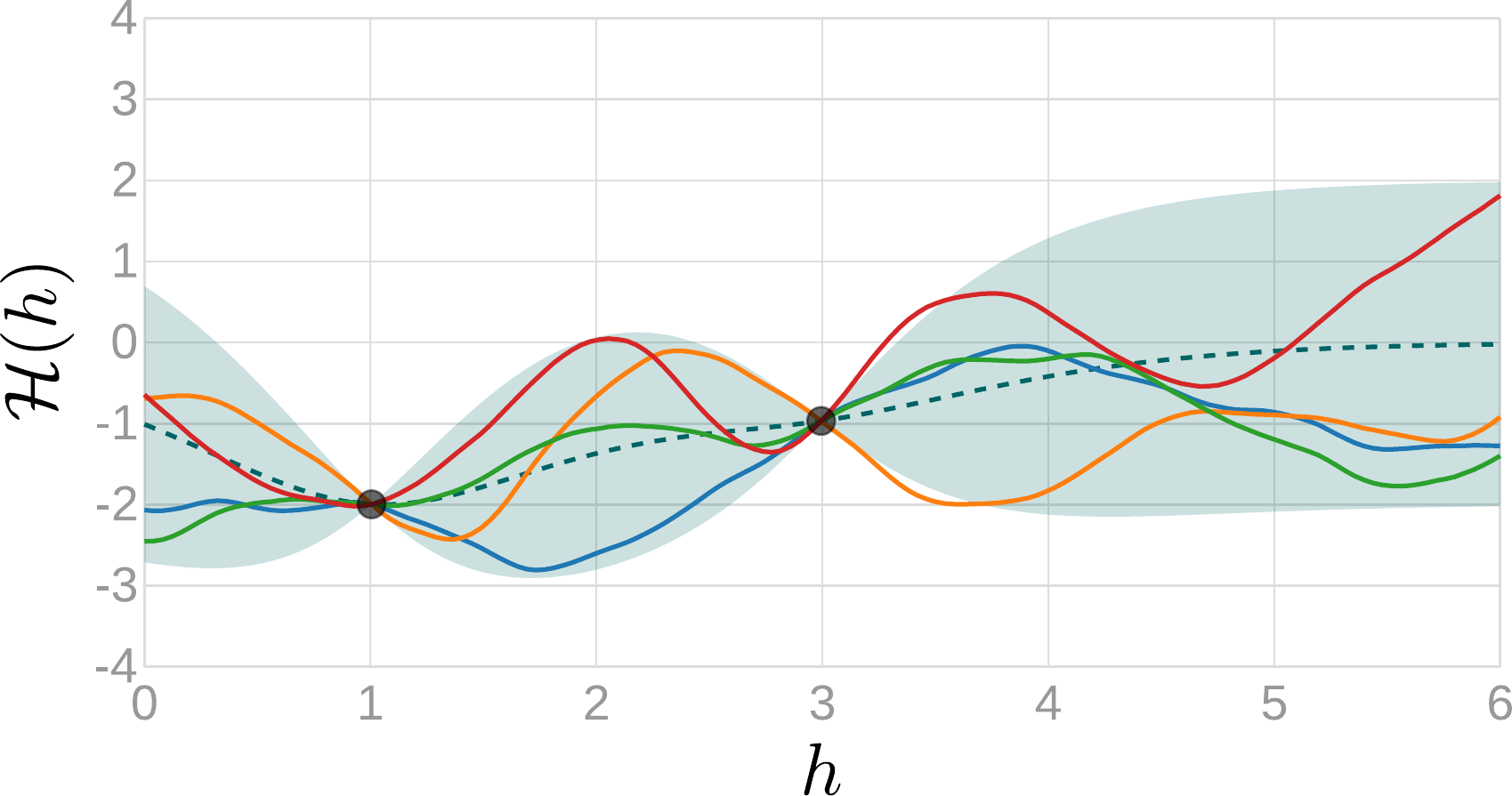}
         \caption{Posterior distribution after two observations.}
         \label{fig:gp_posterior}
     \end{subfigure}
     \begin{subfigure}[t]{0.49\textwidth}
         \centering
         \includegraphics[width=\textwidth]{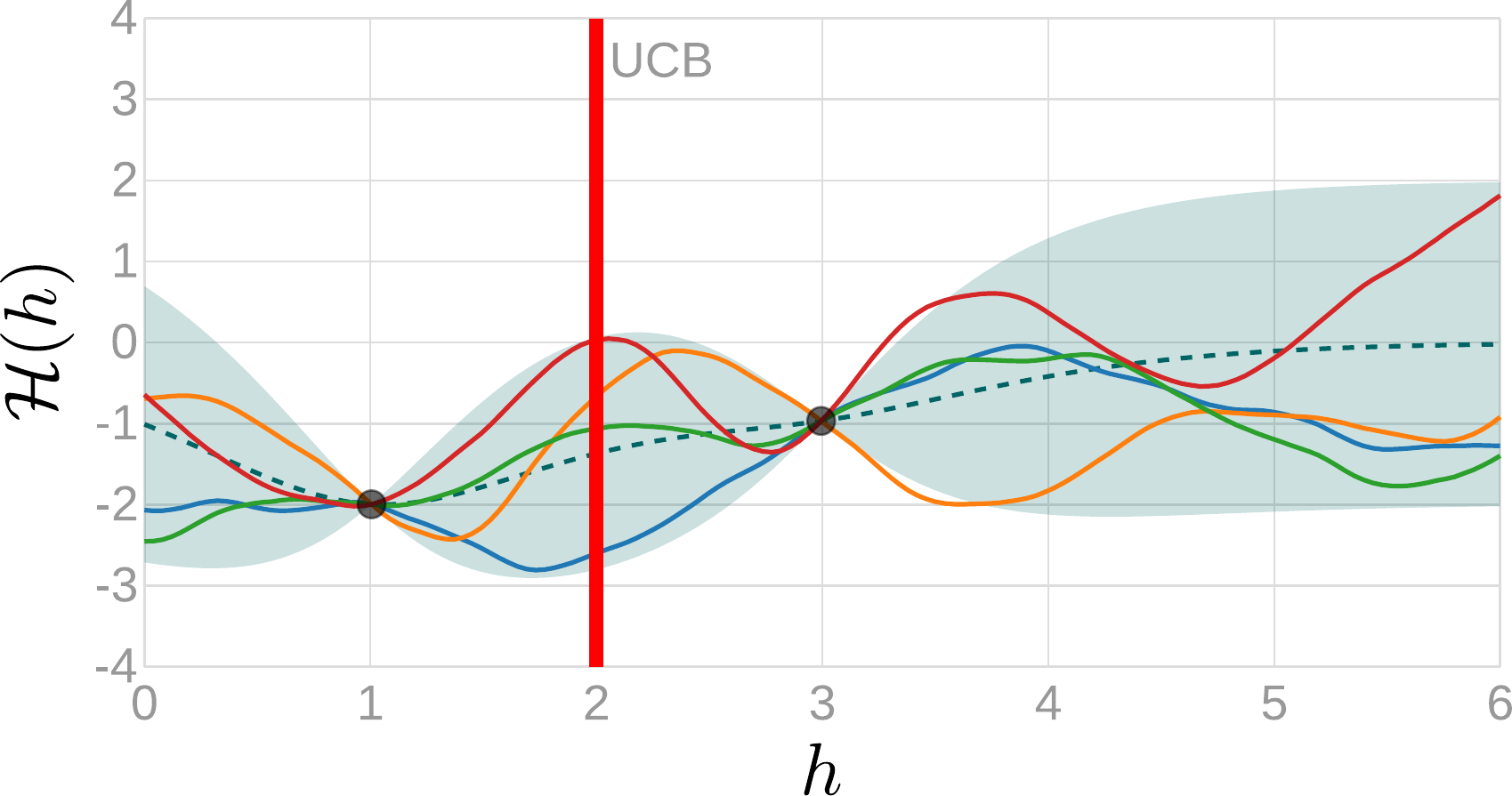}
         \caption{Upper Confidence Bound (UCB) criteria for input acquisition.}
         \label{fig:gp_ucb}
     \end{subfigure}
     \hfill
     \begin{subfigure}[t]{0.49\textwidth}
         \centering
         \includegraphics[width=\textwidth]{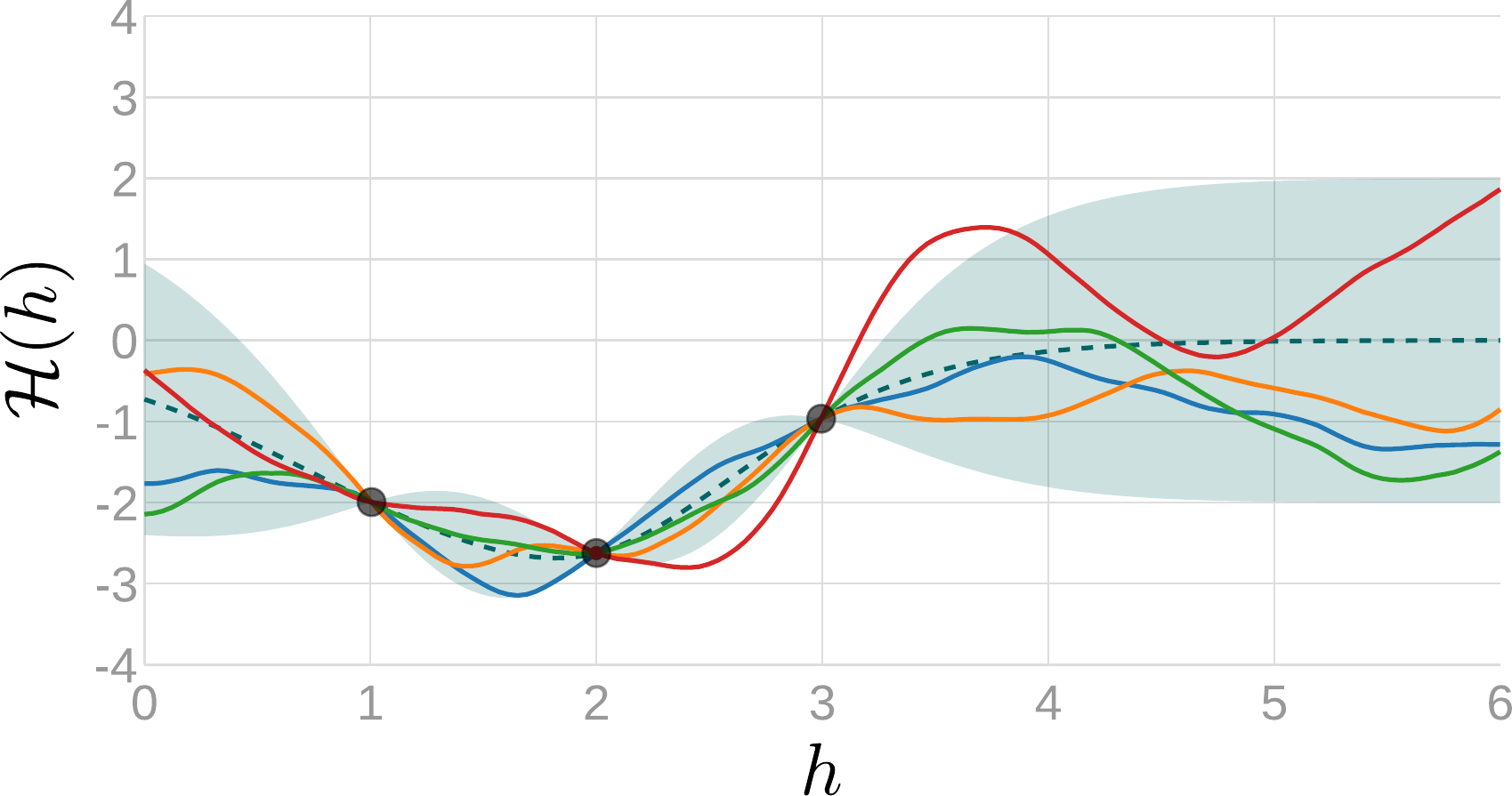}
         \caption{Distribution after acquiring the selected input.}
         \label{fig:gp_final}
     \end{subfigure}
        \caption{Illustration of a GP for fitting and optimizing a scoring function with a single parameter. The dashed line is the expected value of the modeled function, while the colored lines correspond to random sampled functions from the GP distribution. The shaded area represents the 95\% confidence interval at each input value (2$\sigma$).}
        \label{fig:gps}
\end{figure}

\Cref{fig:gps} illustrates the concept for a hypothetical GP with a single input variable $h\in [0, 6]$. We use a zero mean function $m(h)=0$ and a Matérn covariance function~\cite{Rasmussen} with $\nu=5/2$, which is widely used in hyperparameter optimization~\cite{omalley2019kerastuner}. This kernel is defined as follows:

\begin{equation}
k(h, h^\prime) = \left( 1 + \sqrt{5}\|h- h^\prime\|_{l_2}+\frac{5\|h- h^\prime\|_{l_2}^2}{3} \right) * \exp (-\sqrt{5}\|h - h^\prime\|_{l_2}),
\end{equation}

\Cref{fig:gp_prior} shows the prior distribution of the $\mathcal{H}(h)$ function and four random sampled functions following that prior. All samples are relatively smooth (this is determined by the selected kernel), but there is great variability among them. This gives us an idea of the flexibility of the GP in modelling $\mathcal{H}(h)$, but it also shows that, as expected, these priors will not provide useful predictions.

However, as soon as we start measuring some actual values of the scoring function, the model quickly converges to a well-constrained curve. For example, in \Cref{fig:gp_posterior} we incorporate the constraints (measurements) $\mathcal{H}(1)=-2$ and $\mathcal{H}(3)=-1$. Then, the posterior allows us to estimate the value of the scoring function more accurately, especially for the inputs that are closer to the observations.

The calculation of the posterior is based on the assumption that the observations and the desired estimations follow a joint Gaussian distribution. Let us assume we have observed $\mathbf{z}_1 = \mathcal{H}(H_1)$ observations, with $|H_1| = n_1$, and we want to estimate the posterior $\mathbf{z}_2$ for a set $H_2$ of inputs, with $|H_2| = n_2$. Since $\mathbf{z}_1$ and $\mathbf{z}_2$ are jointly Gaussian, we can write:

\begin{equation}
\left[\begin{array}{c} \mathbf{z}_{1} \\ \mathbf{z}_{2} \end{array}\right]
\sim
\mathcal{N} \left(
\left[\begin{array}{c} \mu_{1} \\ \mu_{2} \end{array}\right],
\left[ \begin{array}{cc}
\Sigma_{11} & \Sigma_{12} \\
\Sigma_{21} & \Sigma_{22}
\end{array} \right]
\right)
\end{equation}

with:
\begin{align*}
\mu_{1} & = m(H_1) \quad (n_1 \times 1) \\
\mu_{2} & = m(H_2) \quad (n_2 \times 1) \\
\Sigma_{11} & = k(H_1,H_1) \quad (n_1 \times n_1) \\
\Sigma_{22} & = k(H_2,H_2) \quad (n_2 \times n_2) \\
\Sigma_{12} & = k(H_1,H_2) = \Sigma_{21}^\top \quad (n_1 \times n_2)
\end{align*}

Then, we can calculate the conditional distribution:
\begin{align}
p(\mathbf{z}_2 \mid \mathbf{z}_1) & =  \mathcal{N}(\mu_{2|1}, \Sigma_{2|1}) \label{eq:posterior} \\
\mu_{2|1} & = \mu_2 + \Sigma_{21} \Sigma_{11}^{-1} (\mathbf{z}_1 - \mu_1) \nonumber \\
\Sigma_{2|1} & = \Sigma_{22} - \Sigma_{21} \Sigma_{11}^{-1}\Sigma_{12} \nonumber
\end{align}

In this way, for each $h^*\in H_2$ we can calculate its posterior expected value $\mu_{h^*}$ and standard deviation $\sigma_{h^*}$ according to \Cref{eq:posterior}, as illustrated in \Cref{fig:gp_posterior} for $H_2 = [0, 6]$. It is important to highlight that according to \Cref{eq:posterior}, $p(\mathbf{z}_1 \mid \mathbf{z}_1) \sim \mathcal{N}(\mu_1, 0)$, and therefore it is guaranteed that all the functions taken from the above distribution pass through the observation points.

\paragraph{b) Optimizing the hyperparameter search:} In addition to a probabilistic estimation of the scoring function, the use of GPs as a surrogate model allows us to determine the next set of hyperparameters to test during the optimization. First, we evaluate the scoring function for a small number of random hyperparameter sets, and we construct the initial estimate of the surrogate model. Then, at each iteration, the Bayesian approach consists of the following steps: (1) to select the hyperparameter set to estimate the scoring function; (2) to evaluate the scoring function for the selected hyperparameter set; and (3) to update the surrogate model using the observation obtained in the previous step~\cite{Tipping2004,Theodoridis, FoxRoberts12air}. For step (1), since an explicit formulation of the model is inaccessible, we need to rely on the so-called \textit{acquisition functions}, which estimate an expected loss from evaluating $\mathcal{H}$ at a point $h^*$. Here, we consider the Upper Confidence Bound (UCB)~\cite{Srinivas2012} as our acquisition function, defined as follows:

\begin{equation}
{UCB}_\mathcal{H}(h^*) = \mu_{h^*} - \alpha \sigma_{h^*},
\end{equation}

\noindent where  $\alpha$ is a calibration variable to balance exploration and exploitation. The concept of exploration is here related to the uncertainty derived from the standard deviation $\sigma_{h^*}$. Therefore, the higher the weight assigned to this factor (larger $\alpha$), the more exploratory the behavior of the $UCB$, selecting points further away from those already known. A lower $\alpha$, on the other side, will give more weight to the points observed so far, which corresponds to exploitation. We empirically select $\alpha$ to be $2.6$ \cite{Bayesian, omalley2019kerastuner, DBLP:journals/corr/abs-2104-10201}. \Cref{fig:gp_ucb} shows the minimum $UCB$ value for the model fitted in \Cref{fig:gp_posterior}, and \Cref{fig:gp_final} shows the updated model after incorporating the new scoring function result. Since GPs assume a continuous domain, we need to restrict the $UCB$ and posterior evaluation to those sets of hyperparameters that are actually acceptable for our DNN architecture. With this approach, we aim to optimize the hyperparameters by learning from previous experiments, and hence we expect to require fewer iterations and lower computational time compared to random search.

\section{Numerical results}
\label{sec:results}
\subsection{Hyperparameter tuning}
To increase the speed of the hyperparameter tuning algorithms, we execute them using only 30,000 samples. Then, we train the quasi-optimal DNNs selected by the random search and the Bayesian algorithm using 300,000 samples to find the final DNN approximations of ${\cal F}$ and ${\cal I}$. 

We consider the DNN architecture with hyperparameters $h^o_f = \{n=5, k_0=3, k_1 =3, l = 1\}$ that leads to 525,373 parameters as our reference approximation of ${\cal F}$. Analogously, $h^o_i = \{n=6, k_0=3, k_1 =3\}$ is the set of hyperparameters corresponding to the DNN architecture of the reference DNN approximating ${\cal I}$. The aforementioned DNN architecture consists of 890,925 parameters (see \cite{Shahriari_tool} for more details). To increase the computational efficiency, we also enforce two stopping criteria:
\begin{enumerate}
 \item An early stopping condition with the validation loss variation threshold and the patience being $10^{-3}$ and $30$, respectively. This means that if the change in the loss is below the threshold during $30$ consecutive epochs, the training stops.
 \item If $\mathcal{H}(h) \le 1.1 \times \mathcal{H}(h^o)$, where $h$ is the hyperparameter set under trial, we also stop the training.
\end{enumerate}

\Cref{tab:comp_time_tune} shows the computational time of the hyperparameter tuning using both random search and the Bayesian approach. Results show that the Bayesian approach is less expensive than the random search. \mosir{Notice these time differences will increase as we augment the number of unknowns (i.e., measurements in the forward problem, and inverted parameters in the inverse problem).}

\begin{figure}[ht]
	\centering
	\begin{tikzpicture}
	\node at (0,0.0)[scale=1]{
\begin{tikzpicture}
\begin{axis}[
legend columns=2,
height=0.35*\textwidth,
width=1*\textwidth,
xlabel={Number of trainable parameters},
ylabel near ticks,
ylabel={Score},
]

\addplot[only marks,scatter,mark=otimes*,mark size=1.5pt] table [x=Trainable_Parameters, y=Score]{results/automl_new_loss/without_max/Forward_bayesian_score_full.txt};


\end{axis}	
\end{tikzpicture}};
	\node (C_1) at (0,6.7) {Individual scoring components};
	\node at (0,4.5)[scale=1]{
\begin{tikzpicture}
\begin{axis}[
legend columns=2,
height=0.35*\textwidth,
width=\textwidth,
xlabel={Number of trainable parameters},
ylabel near ticks,
ylabel={Score},
legend pos=north east,
legend columns=1,
]

\addplot[thick] table [x=Trainable_Parameters, y=Score_unknowns]{results/automl_new_loss/without_max/Forward_bayesian_score_full.txt};

\addlegendentry{\scriptsize Relative increase in the number of unknowns}

\addplot[only marks,scatter,mark=otimes*,mark size=1.5pt] table [x=Trainable_Parameters, y=Score_loss]{results/automl_new_loss/without_max/Forward_bayesian_score_full.txt};

\addlegendentry{\scriptsize Relative error}

\end{axis}	
\end{tikzpicture}};
	\node (C_2) at (0,2.2) {Overall score};
	\end{tikzpicture}
	\caption{ DNN optimization of the forward function ${\cal F}$ using a Bayesian approach. The colors indicate separate clusters of points.}
	\label{fig:forward_opt_bayesian}
\end{figure}
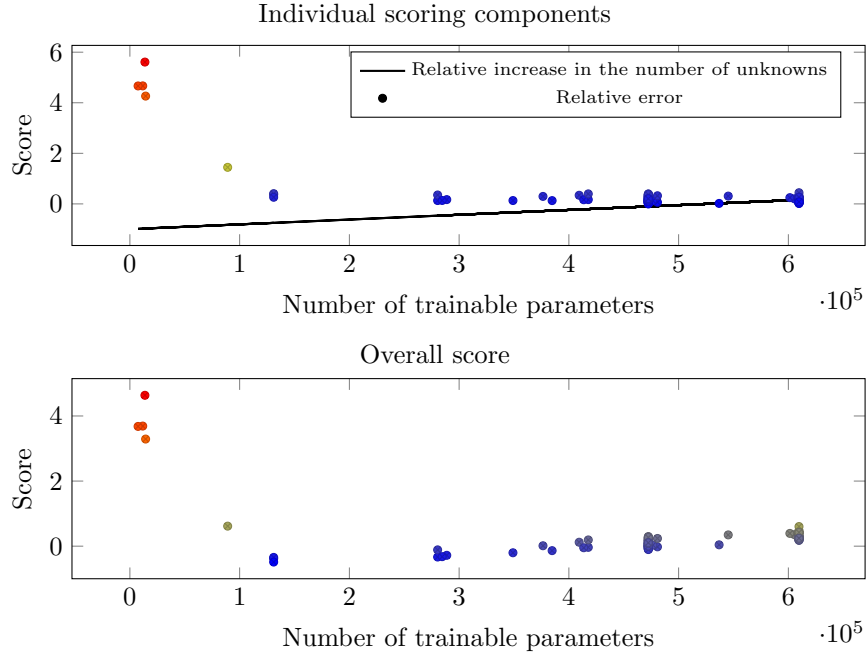

\begin{figure}[ht]
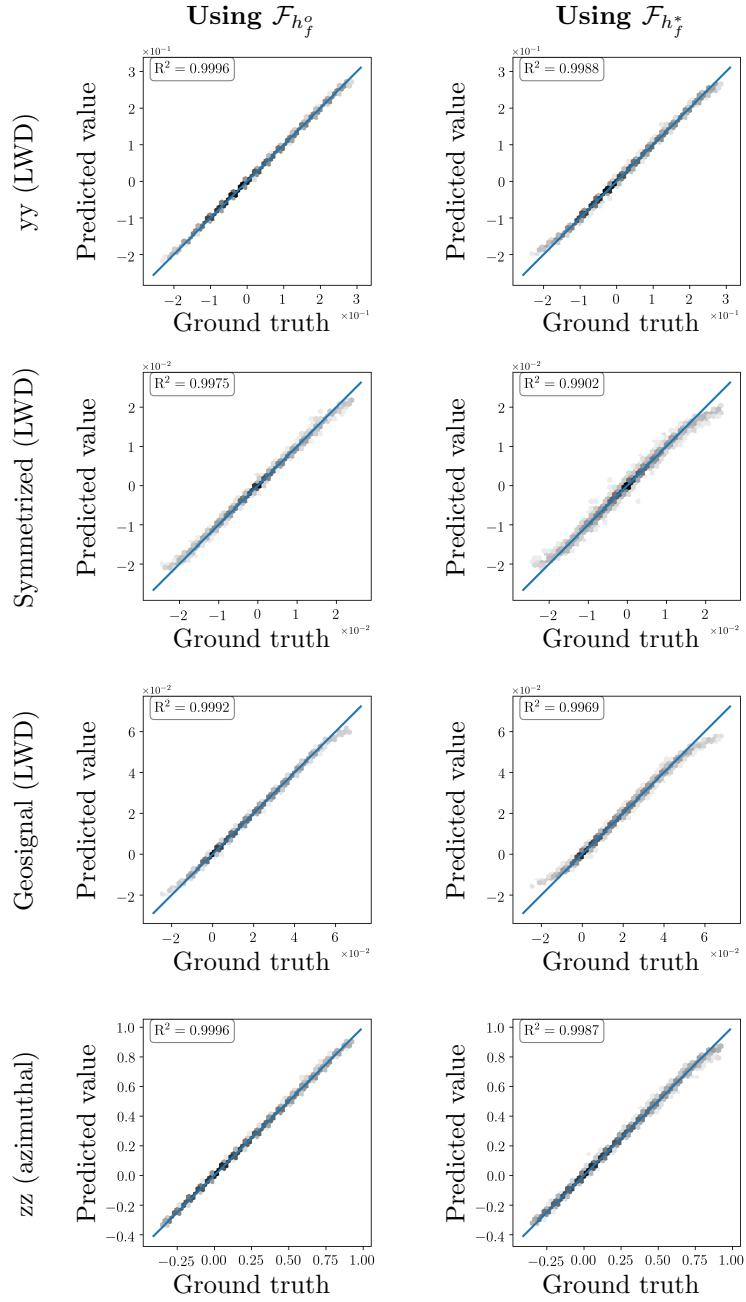

	\begin{tikzpicture}
	\node at (0,0.0)[scale=1]{\input{results/forward_original/atten_7.tex}};
	
	\node at (5,0.0)[scale=1]{\input{results/forward_optimal/atten_7.tex}};
	
	\node at (0,-4.3)[scale=1]{\input{results/forward_original/atten_3.tex}};
	
	\node at (5,-4.3)[scale=1]{\input{results/forward_optimal/atten_3.tex}};

	%
	\node at (0,-8.6)[scale=1]{\input{results/forward_original/atten_4.tex}};
	
	\node at (5,-8.6)[scale=1]{\input{results/forward_optimal/atten_4.tex}};
	
	\node at (0,-12.9)[scale=1]{\input{results/forward_original/atten_5.tex}};
	
	\node at (5,-12.9)[scale=1]{\input{results/forward_optimal/atten_5.tex}};
	
	\end{tikzpicture}
	
	\caption{DNN approximation of ${\cal F}$. Comparison between the original network and the quasi-optimal one for the real part of selected measurements.}
	\label{fig:measurements}
\end{figure}

\mosinn{\Cref{fig:forward_opt_randomsearch} shows the results of hyperparameter tuning using random search to approximate ${\cal F}$. It represents the score value for each selection of hyperparameters from the search space ${\cal S}_{\cal F}$ and its corresponding number of trainable parameters. We also display the effect of individual factors involved in the scoring function, i.e., the relative decrease in the number of unknowns and the relative error. Analogously, \Cref{fig:forward_opt_bayesian} shows the results of tuning using the Bayesian approach to approximate ${\cal F}$. In the case of random search, the algorithm repeatedly considers DNN architectures with less than 100,000 parameters even when their score is not significantly improving. However, in the Bayesian approach, the algorithm rapidly learns that this cluster of DNN architectures leads to unacceptable scores. Considering the results of both algorithms, we witness that the quasi-optimal set of hyperparameters is $h_f^{\ast} = \{n=3, k_0=3, k_1 =3, l = 7\}$, which corresponds to 131,013 parameters. Using this DNN, we achieve a comparable score to the reference one with approximately $25\%$ of the trainable parameters used by the reference DNN. \Cref{fig:measurements} shows cross-plots comparing the accuracy of the DNN approximation of ${\cal F}$ using the quasi-optimal DNN and the reference one for a selected set of measurements. The accuracy of the two DNNs is similar, i.e., the $R^2$ scores of the prediction vs. ground truth are comparable. Moreover, according to the training time shown in \Cref{tab:comp_time}, training the reference DNN takes almost four times more computational time compared to the quasi-optimal one.}

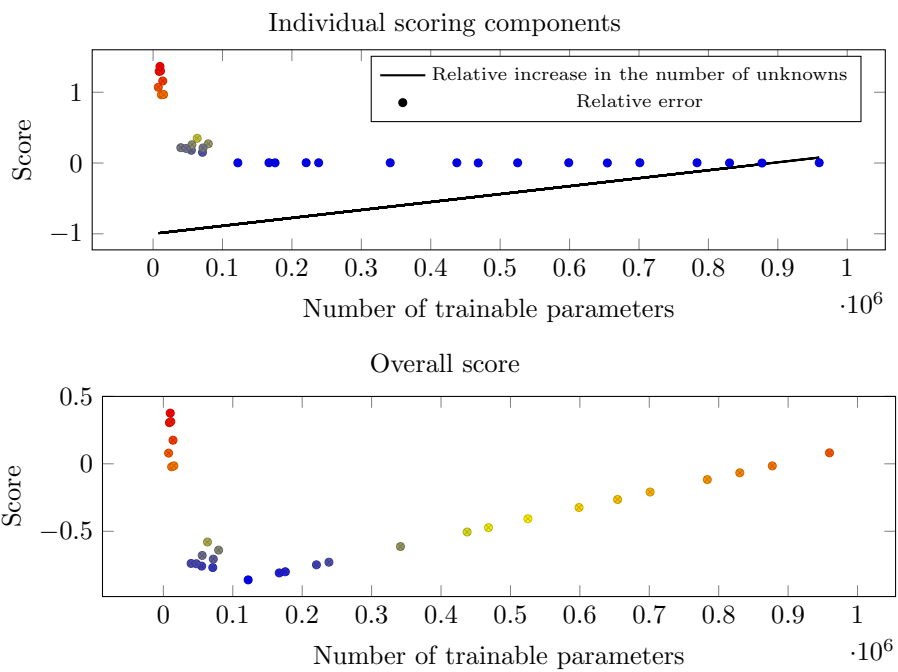
\begin{figure}[ht]
	\centering
	\begin{tikzpicture}
	\node at (0,0.0)[scale=1]{
\begin{tikzpicture}
\begin{axis}[
legend columns=2,
height=0.35*\textwidth,
width=1*\textwidth,
xlabel={Number of trainable parameters},
ylabel near ticks,
ylabel={Score},
]

\addplot[only marks,scatter,mark=otimes*,mark size=1.5pt] table [x=Trainable_Parameters, y=Score]{results/automl_new_loss/without_max/Inverse_random_search_score_full.txt};


\end{axis}	
\end{tikzpicture}};
	\node (C_1) at (0,6.7) {Individual scoring components};
	\node at (0,4.5)[scale=1]{
\begin{tikzpicture}
\begin{axis}[
legend columns=2,
height=0.35*\textwidth,
width=\textwidth,
xlabel={Number of trainable parameters},
ylabel near ticks,
ylabel={Score},
legend pos=north east,
legend columns=1,
]

\addplot[thick] table [x=Trainable_Parameters, y=Score_unknowns]{results/automl_new_loss/without_max/Inverse_random_search_score_full.txt};

\addlegendentry{\scriptsize Relative increase in the number of unknowns}

\addplot[only marks,scatter,mark=otimes*,mark size=1.5pt] table [x=Trainable_Parameters, y=Score_loss]{results/automl_new_loss/without_max/Inverse_random_search_score_full.txt};

\addlegendentry{\scriptsize Relative error}

\end{axis}	
\end{tikzpicture}};
	\node (C_2) at (0,2.2) {Overall score};
	\end{tikzpicture}
	\caption{DNN optimization of the inverse function ${\cal I}$ using a random search. The colors indicate separate clusters of points.}
	\label{fig:inverse_opt_randomsearch}
\end{figure}

\begin{figure}[ht]
	\centering
	\begin{tikzpicture}
	\node at (0,0.0)[scale=1]{
\begin{tikzpicture}
\begin{axis}[
legend columns=2,
height=0.35*\textwidth,
width=1*\textwidth,
xlabel={Number of trainable parameters},
ylabel near ticks,
ylabel={Score},
]

\addplot[only marks,scatter,mark=otimes*,mark size=1.5pt] table [x=Trainable_Parameters, y=Score]{results/automl_new_loss/without_max/Inverse_bayesian_score_full.txt};


\end{axis}	
\end{tikzpicture}};
	\node (C_1) at (0,6.7) {Individual scoring components};
	\node at (0,4.5)[scale=1]{
\begin{tikzpicture}
\begin{axis}[
legend columns=2,
height=0.35*\textwidth,
width=\textwidth,
xlabel={Number of trainable parameters},
ylabel near ticks,
ylabel={Score},
legend pos=north east,
legend columns=1,
]

\addplot[thick] table [x=Trainable_Parameters, y=Score_unknowns]{results/automl_new_loss/without_max/Inverse_bayesian_score_full.txt};

\addlegendentry{\scriptsize Relative increase in the number of unknowns}

\addplot[only marks,scatter,mark=otimes*,mark size=1.5pt] table [x=Trainable_Parameters, y=Score_loss]{results/automl_new_loss/without_max/Inverse_bayesian_score_full.txt};

\addlegendentry{\scriptsize Relative error}

\end{axis}	
\end{tikzpicture}};
	\node (C_2) at (0,2.2) {Overall score};
	\end{tikzpicture}
	\caption{ DNN optimization of the inverse function ${\cal I}$ using a Bayesian approach. The colors indicate separate clusters of points.}
	\label{fig:inverse_opt_bayesian}
\end{figure}
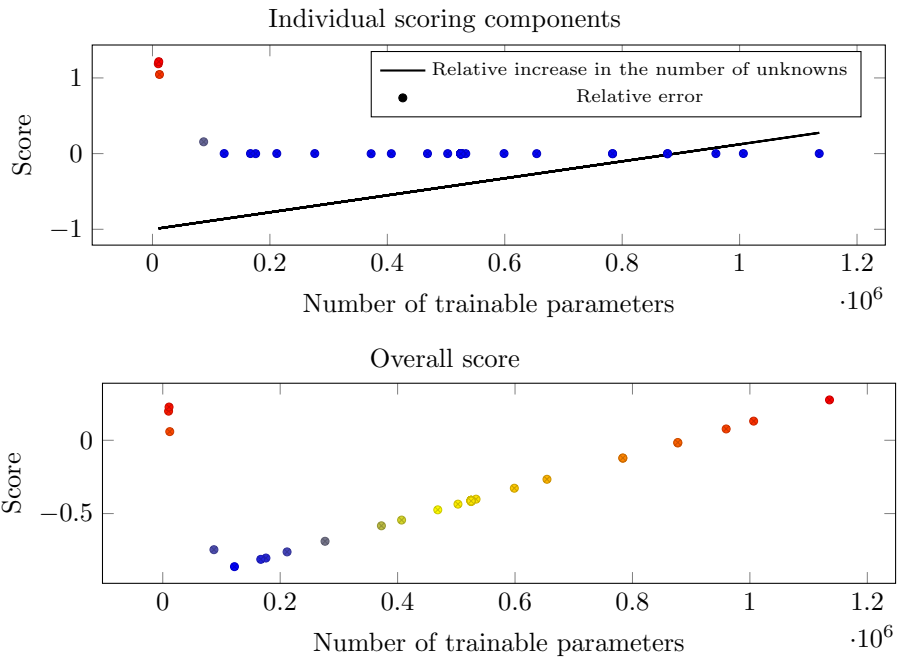

\begin{figure}[ht]
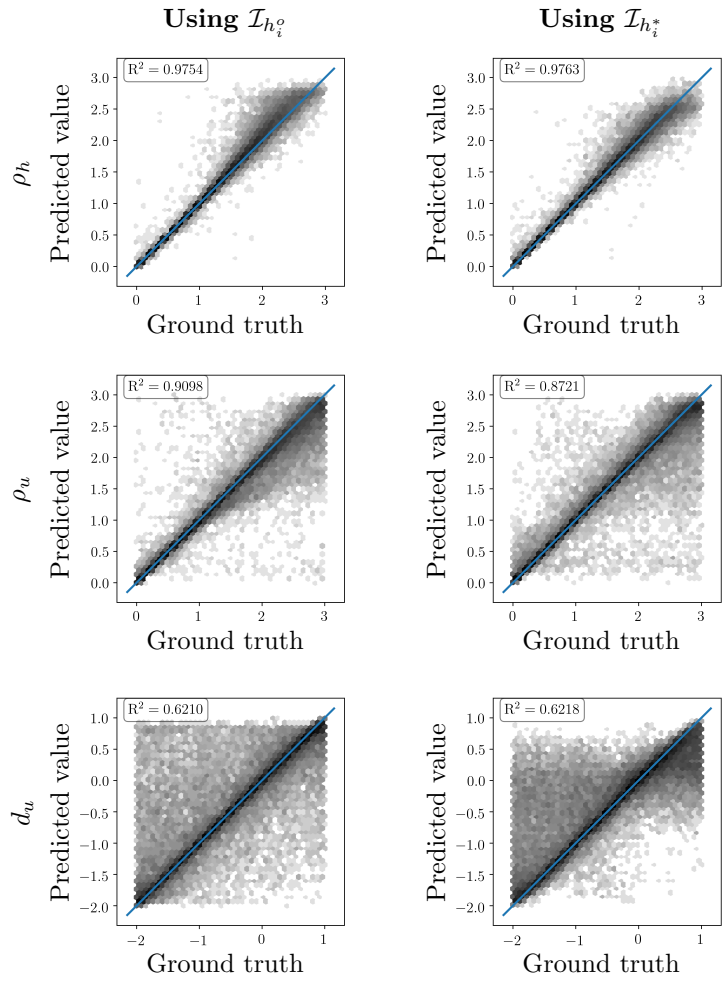

	\begin{tikzpicture}
	\node at (0,0.0)[scale=1]{\input{results/inverse_original/rho_h.tex}};
	
	\node at (5,0.0)[scale=1]{\input{results/inverse_optimal/rho_h.tex}};
	
	\node at (0,-4.3)[scale=1]{\input{results/inverse_original/rho_u.tex}};
	
	\node at (5,-4.3)[scale=1]{\input{results/inverse_optimal/rho_u.tex}};

	%
	\node at (0,-8.6)[scale=1]{\input{results/inverse_original/d_u.tex}};
	
	\node at (5,-8.6)[scale=1]{\input{results/inverse_optimal/d_u.tex}};

	\end{tikzpicture}
	
	\caption{DNN approximation of ${\cal I}$. Comparison between the original network and the quasi-optimal one for a selected set of material properties.}
	\label{fig:variables}
\end{figure}

\Cref{fig:inverse_opt_randomsearch} and \Cref{fig:inverse_opt_bayesian} show the process of hyperparameter tuning to obtain a quasi-optimal DNN architecture to approximate ${\cal I}$. Analogous to the tuning for the forward function, the Bayesian approach selects a less redundant set of hyperparameters compared to the random search. By comparing the scores of all the DNN architectures, the quasi-optimal set of hyperparameters is $h_i^{\ast} = \{n=3, k_0=3, k_1 =3\}$ with 122,125 parameters. The quasi-optimal DNN architecture contains more than seven times fewer parameters than the reference one. \Cref{fig:variables} compares the accuracy of the quasi-optimal DNN architecture and the reference one for some inversion variables. \Cref{tab:comp_time} shows that we spend almost eight times more computational time to train the reference DNN compared to the quasi-optimal one.

\begin{table}[ht]
	\centering
	\begin{tabular}{r|c|c}
		\toprule
		problem   &  Random search [\si{\hour}] & Bayesian approach [\si{\hour}]   \\
		\midrule
		Forward & \num{18.02} & \num{16.3} \\
		Inverse    & \num{5.97} & \num{4.80} \\
		\bottomrule
	\end{tabular}
	\caption{Comparison of the time required to perform a random search vs a Bayesian approach.}
	\label{tab:comp_time_tune}
\end{table}

\begin{figure}[ht]
	\centering
	\begin{tikzpicture}
	\node at (0,0.0)[scale=1]{
\begin{tikzpicture}
\begin{axis}[
legend columns=2,
height=0.35*\textwidth,
width=1*\textwidth,
xlabel={Number of trainable parameters},
ylabel near ticks,
ylabel={Score},
]

\addplot[only marks,scatter,mark=otimes*,mark size=1.5pt] table [x=Trainable_Parameters, y=Score]{results/automl_new_loss/without_max/Forward_random_search_score_full.txt};

\end{axis}	
\end{tikzpicture}};
	\node (C_1) at (0,6.7) {Individual scoring components};
	\node at (0,4.5)[scale=1]{
\begin{tikzpicture}
\begin{axis}[
legend columns=2,
height=0.35*\textwidth,
width=\textwidth,
xlabel={Number of trainable parameters},
ylabel near ticks,
ylabel={Score},
legend pos=north east,
legend columns=1,
]

\addplot[thick] table [x=Trainable_Parameters, y=Score_unknowns]{results/automl_new_loss/without_max/Forward_random_search_score_full.txt};

\addlegendentry{\scriptsize Relative increase in the number of unknowns}

\addplot[only marks,scatter,mark=otimes*,mark size=1.5pt] table [x=Trainable_Parameters, y=Score_loss]{results/automl_new_loss/without_max/Forward_random_search_score_full.txt};

\addlegendentry{\scriptsize Relative error}

\end{axis}	
\end{tikzpicture}};
	\node (C_2) at (0,2.2) {Overall score};
	\end{tikzpicture}
	\caption{DNN optimization of the forward function ${\cal F}$ using a random search. The colors indicate separate clusters of points.}
	\label{fig:forward_opt_randomsearch}
\end{figure}
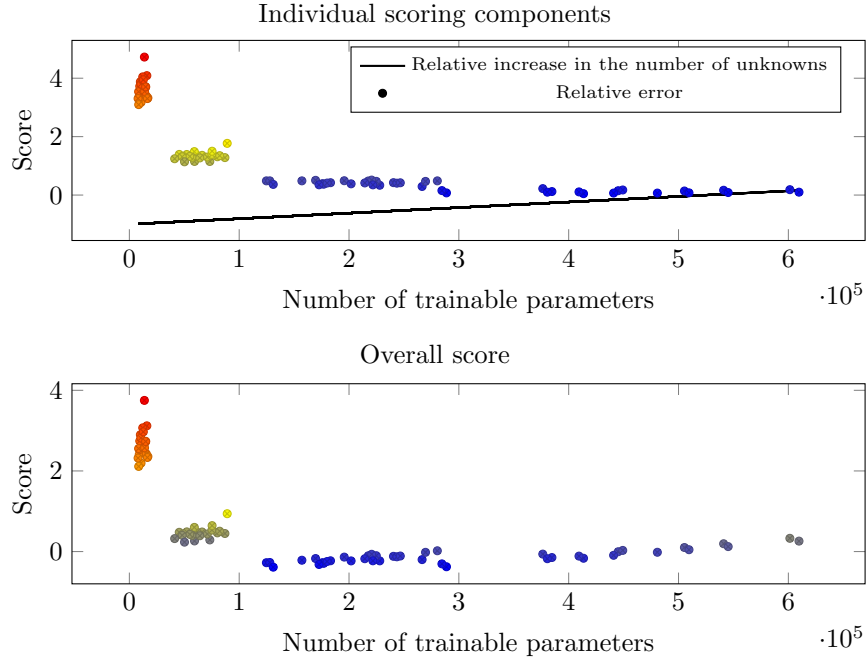

\begin{table}[ht]
	\centering
	\begin{tabular}{r|c|c}
		\toprule
		problem   &  original DNN training time [\si{\hour}] & quasi-optimal DNN training time [\si{\hour}]   \\
		\midrule
		Forward & \num{18.69} & \num{4.65} \\
		Inverse    & \num{34.41} & \num{4.16} \\
		\bottomrule
	\end{tabular}
	\caption{Comparison of the training time between the original DNN and the quasi-optimal one.}
	\label{tab:comp_time}
\end{table}

\subsection{Synthetic example}
\begin{figure}[ht]
	\centering
	\begin{tikzpicture}
	\node at (0,0.0)[scale=1]{\pgfplotsset{every axis legend/.append style={
		at={(0.5,1.03)},
		anchor=south},
	every axis plot/.append style={line width=1.8pt},
}
\begin{tikzpicture}
\begin{axis}[
xmin=0,
xmax=540,
legend columns=2,
ymin=0.0,
 ymax=15,
height=0.30\textwidth,
width=0.70\textwidth,
 y dir=reverse,
xlabel={HD ($m$)},
ylabel near ticks,
ylabel={TVD ($m$)},
enlargelimits=false,
]

\addplot graphics[xmin=0,xmax=540,ymin=0,ymax=15] {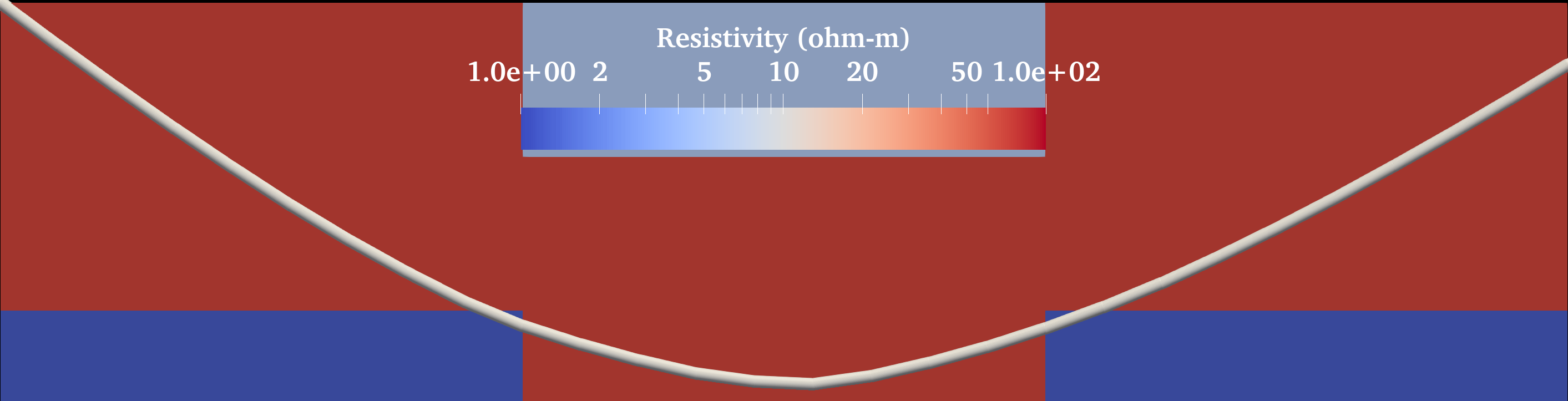};

\end{axis}	

\node[rotate=90] (I_h) at (-1.50,1) {Synthetic formation};
\end{tikzpicture}};
	\node at (0,-3.5)[scale=1]{\pgfplotsset{every axis legend/.append style={
		at={(0.5,1.03)},
		anchor=south},
	every axis plot/.append style={line width=1.8pt},
}
\begin{tikzpicture}
\begin{axis}[
xmin=0,
xmax=540,
legend columns=2,
ymin=0.0,
 ymax=15,
height=0.30\textwidth,
width=0.70\textwidth,
 y dir=reverse,
xlabel={HD ($m$)},
ylabel near ticks,
ylabel={TVD ($m$)},
enlargelimits=false,
]

\addplot graphics[xmin=0,xmax=540,ymin=0,ymax=15] {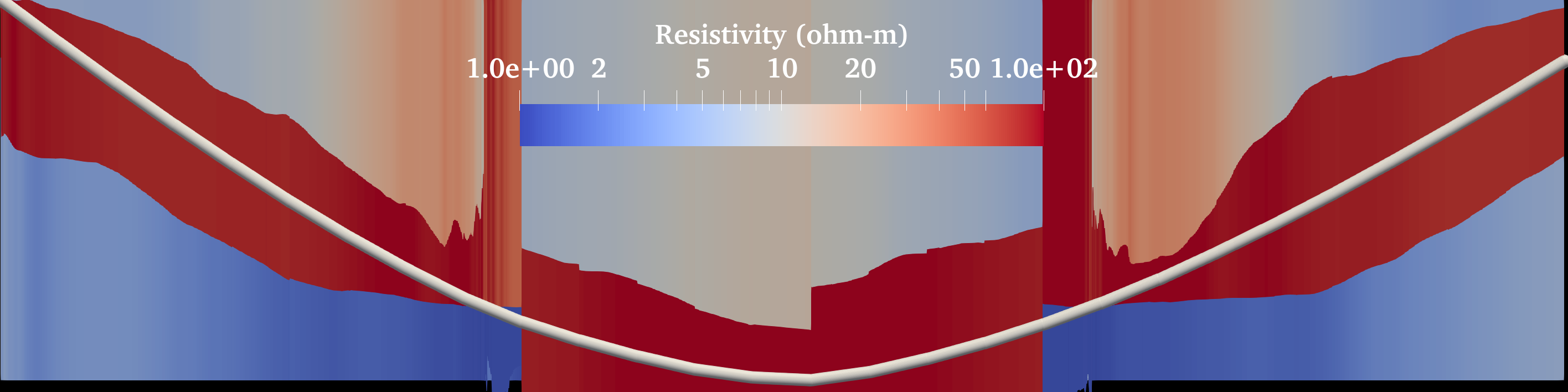};

\end{axis}	

\node[rotate=90] (I_h) at (-1.50,1) {Using ${\cal I}_{h^o_i}$};
\end{tikzpicture}};
		\node at (0,-7)[scale=1]{\pgfplotsset{every axis legend/.append style={
		at={(0.5,1.03)},
		anchor=south},
	every axis plot/.append style={line width=1.8pt},
}
\begin{tikzpicture}
\begin{axis}[
xmin=0,
xmax=540,
legend columns=2,
ymin=0.0,
 ymax=15,
height=0.30\textwidth,
width=0.70\textwidth,
 y dir=reverse,
xlabel={HD ($m$)},
ylabel near ticks,
ylabel={TVD ($m$)},
enlargelimits=false,
]

\addplot graphics[xmin=0,xmax=540,ymin=0,ymax=15] {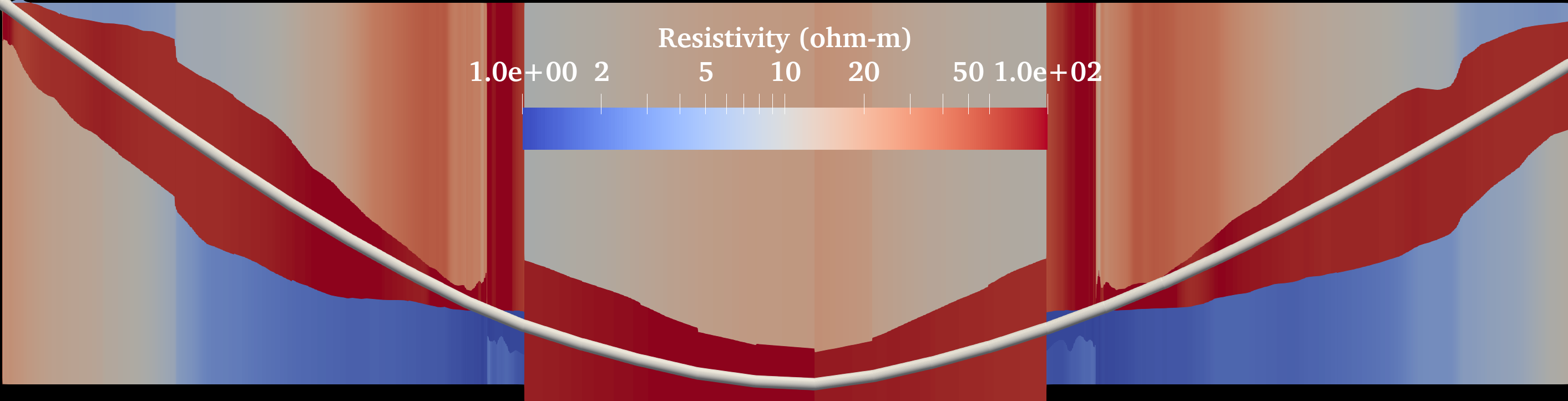};

\end{axis}	

\node[rotate=90] (I_h) at (-1.50,1) {Using ${\cal I}_{h^\ast_i}$};
\end{tikzpicture}};
	\end{tikzpicture}
	\caption{ Model problem 1. Comparison amongst the synthetic (original) formation, and the formations predicted by the original (reference) DNN, and the quasi-optimal DNN.}
	\label{fig:problem_1}
\end{figure}

\begin{figure}[ht]
	\centering
	\begin{tikzpicture}
	\node at (0,0.0)[scale=1]{\pgfplotsset{every axis legend/.append style={
		at={(0.5,1.03)},
		anchor=south},
	every axis plot/.append style={line width=1.8pt},
}
\begin{tikzpicture}
\begin{axis}[
xmin=0,
xmax=540,
legend columns=2,
ymin=0.0,
 ymax=15,
height=0.30\textwidth,
width=0.70\textwidth,
 y dir=reverse,
xlabel={HD ($m$)},
ylabel near ticks,
ylabel={TVD ($m$)},
enlargelimits=false,
]

\addplot graphics[xmin=0,xmax=540,ymin=0,ymax=15] {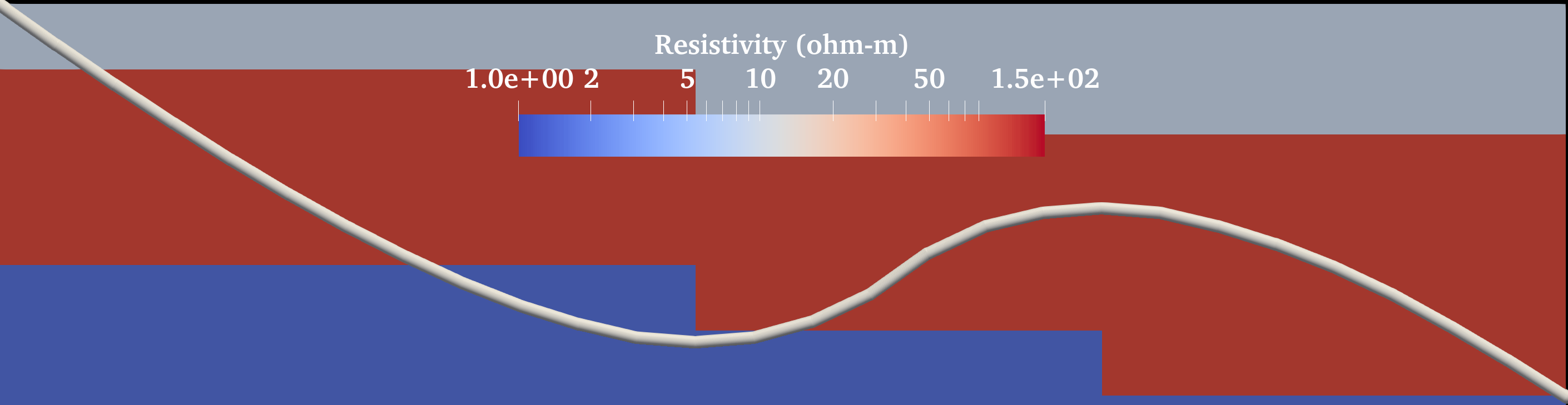};

\end{axis}	

\node[rotate=90] (I_h) at (-1.50,1) {Synthetic formation};
\end{tikzpicture}};
	\node at (0,-3.5)[scale=1]{\pgfplotsset{every axis legend/.append style={
		at={(0.5,1.03)},
		anchor=south},
	every axis plot/.append style={line width=1.8pt},
}
\begin{tikzpicture}
\begin{axis}[
xmin=0,
xmax=540,
legend columns=2,
ymin=0.0,
 ymax=15,
height=0.30\textwidth,
width=0.70\textwidth,
 y dir=reverse,
xlabel={HD ($m$)},
ylabel near ticks,
ylabel={TVD ($m$)},
enlargelimits=false,
]

\addplot graphics[xmin=0,xmax=540,ymin=0,ymax=15] {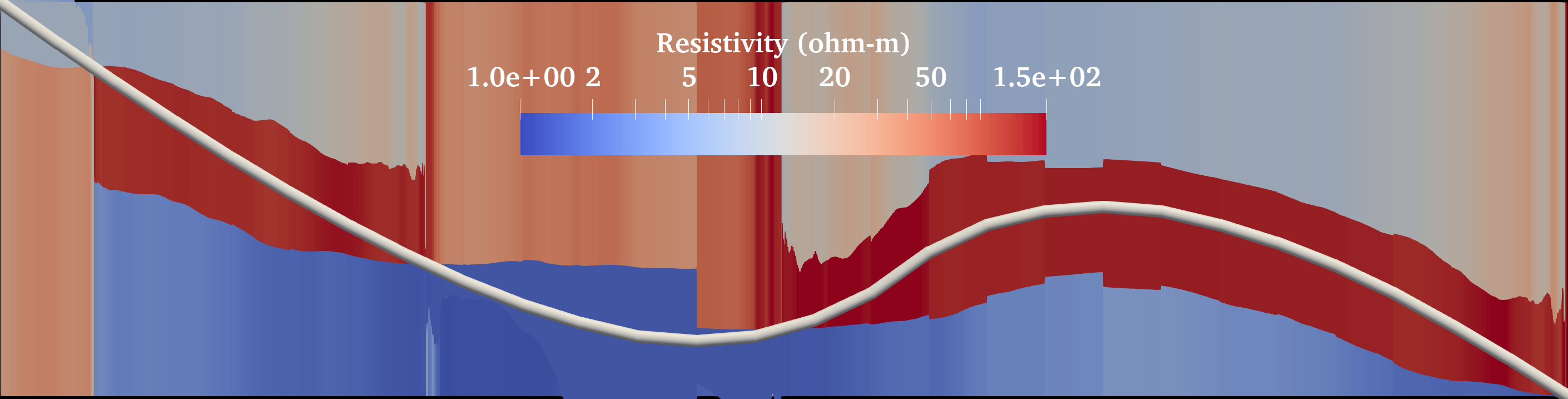};

\end{axis}	

\node[rotate=90] (I_h) at (-1.50,1) {Using ${\cal I}_{h^o_i}$};
\end{tikzpicture}};
	\node at (0,-7)[scale=1]{\pgfplotsset{every axis legend/.append style={
		at={(0.5,1.03)},
		anchor=south},
	every axis plot/.append style={line width=1.8pt},
}
\begin{tikzpicture}
\begin{axis}[
xmin=0,
xmax=540,
legend columns=2,
ymin=0.0,
 ymax=15,
height=0.30\textwidth,
width=0.70\textwidth,
 y dir=reverse,
xlabel={HD ($m$)},
ylabel near ticks,
ylabel={TVD ($m$)},
enlargelimits=false,
]

\addplot graphics[xmin=0,xmax=540,ymin=0,ymax=15] {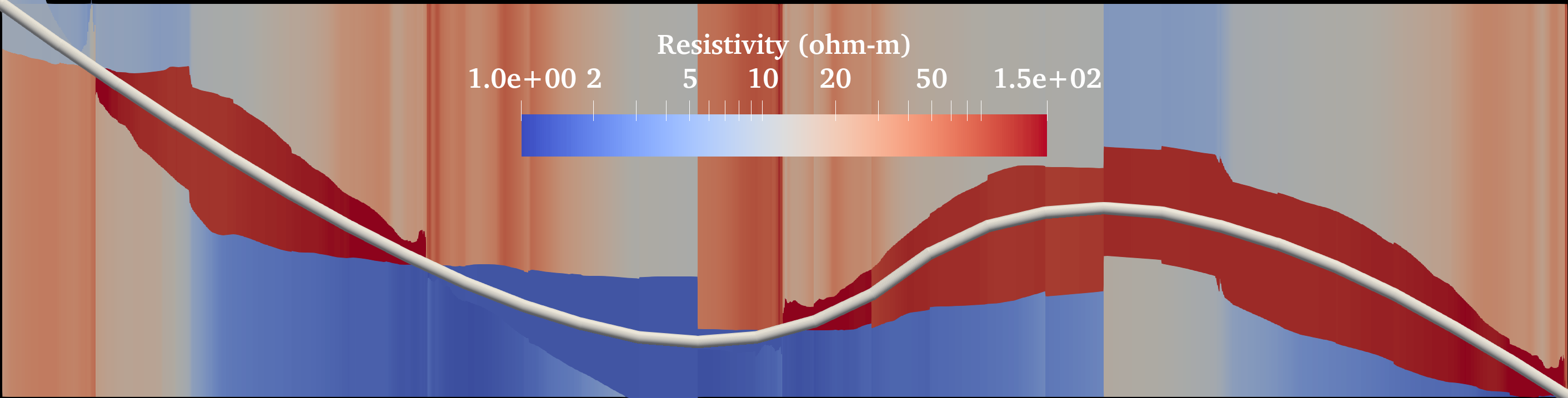};

\end{axis}	

\node[rotate=90] (I_h) at (-1.50,1) {Using ${\cal I}_{h^\ast_i}$};
\end{tikzpicture}};
	\end{tikzpicture}
	\caption{ Model problem 2. Comparison amongst the synthetic (original) formation, and the formations predicted by the original (reference) DNN, and the quasi-optimal DNN.}
	\label{fig:problem_2}
\end{figure}

\Cref{fig:problem_1} compares the inversion results using ${\cal I}_{h^\ast_i}$ and ${\cal I}_{h^o_i}$ to the actual formation for a synthetic model. \mosir{Both inversion models can adequately predict the material properties up to a sufficient depth of investigation. The quasi-optimal DNN predicts the material properties around the trajectory. Moreover, it detects the bed boundary corresponding to oil-to-water contact from a few meters away from the trajectory.} \Cref{fig:problem_2} shows similar results for a second synthetic formation.

\section{Conclusions}
\label{sec:conclusion}
\mosir{In this work, we used AutoML--specifically, DNN architecture search algorithms--to obtain quasi-optimal DNN architectures for the inversion of borehole resistivity measurements. A quasi-optimal DNN provides accurate results with a minimum number of unknowns (trainable parameters). We introduced a scoring function that accounts both for the accuracy of the trained DNN and its size compared to a reference large DNN. We introduced convolutional blocks as the main components of the DNN architecture.}

\mosir{We used two standard search algorithms to find our quasi-optimal hyperparameters: random search and a Bayesian approach based on Gaussian Processes. Both automatic search algorithms deliver quasi-optimal DNN architectures with reduced hand-design. Random search performs an arbitrary selection of the hyperparameters. In contrast, the Bayesian approach purposefully selects the hyperparameters using the information obtained from the previous iterations. Thus, it performs a less redundant selection of hyperparameters, and it typically requires fewer iterations to achieve the quasi-optimal DNN architecture, thereby, requiring less computational time than random search.}
	
\mosir{In this work, both algorithms converged to the same architecture because the search space is relatively small, and we imposed no stopping criteria while searching for the quasi-optimal DNN (tuning). Although the quasi-optimal DNN architecture contains significantly fewer trainable parameters, it still delivers a performance comparable to the original DNN. Moreover, it substantially reduces the computational time required to train the DNN.}

\mosir{In future work, we shall investigate the effect of using noisy data for training and evaluating the DNN. Moreover, we shall consider more complicated scenarios, e.g., a more general two- and three-dimensional subsurface parametrization, possibly combined with transfer learning. In addition, we shall study the possibility of an automated approach based on active learning to efficiently sample the space of subsurface properties using the minimum number of samples, i.e., the minimum dataset's size.}

\section*{Acknowledgments}
Mostafa Shahriari and Somayeh Kargaran have been supported by the Austrian Ministry for Transport, Innovation and Technology (BMVIT), the Federal Ministry for Digital and Economic Affairs (BMDW), the Province of Upper Austria in the frame of the COMET - Competence Centers for Excellent Technologies Program managed by Austrian Research Promotion Agency FFG, the COMET Module S3AI managed by the Austrian Research Promotion Agency FFG, and the ‘‘Austrian COMET-Programme’’ (Project InTribology, no. 872176). 

David Pardo has received funding from: the European Union's Horizon 2020 research and innovation program under the Marie Sklodowska-Curie grant agreement No 777778 (MATHROCKS); the European Regional Development Fund (ERDF) through the Interreg V-A Spain-France-Andorra program POCTEFA 2014-2020 Project PIXIL (EFA362/19); the Spanish Ministry of Science and Innovation projects with references PID2019-108111RB-I00 (FEDER/AEI) and PDC2021-121093-I00 (AEI/Next Generation EU), the “BCAM Severo Ochoa” accreditation of excellence (SEV-2017-0718); and the Basque Government through the BERC 2022-2025 program, the three Elkartek projects 3KIA (KK-2020/00049), EXPERTIA (KK-2021/00048), and SIGZE (KK-2021/00095), and the Consolidated Research Group MATHMODE (IT1456-22) given by the Department of Education.

Tomas Teijeiro is supported by a Maria Zambrano fellowship (MAZAM21/29) from the University of Basque Country and the Spanish Ministry of Universities, funded by the European Union-Next-GenerationEU.
\bibliographystyle{unsrt}
\bibliography{mybibfile}

\end{document}